\pdfoutput=1
\documentclass[10pt,letterpaper]{article}

\usepackage{txbench-preprint}
\definecolor{todotext}{HTML}{A95F3D}

\preprintKicker{bioRxiv preprint}
\preprintCategory{Computational Biology}
\preprintDate{June 2026}
\preprintTitle{TxBench-PP: Analyzing AI Agent Performance on Small-Molecule Preclinical Pharmacology}
\preprintAuthors{Hannah Le\preprintSup{1,*}, Ramesh Ramasamy\preprintSup{1,*}, Alex Urrutia\preprintSup{1,*}, Mahsa Yazdani\preprintSup{1,*}, Tim Proctor\preprintSup{1}, Kenny Workman\preprintSup{1}}
\preprintAffiliations{\preprintSup{1}LatchBio, San Francisco, CA, USA\\
\preprintSup{*}These authors contributed equally.}
\preprintCorrespondence{kenny@latch.bio}

\begin{document}

\PreprintHeader

\begin{PreprintAbstract}
Artificial intelligence (AI) agents promise to accelerate drug discovery by
compressing interpretation and decision-making loops, but practical deployment
requires trusted evaluation on realistic program decisions. We introduce
TherapeuticsBench Preclinical Pharmacology (TxBench-PP), a verifiable benchmark
for small-molecule
preclinical pharmacology and the first focused slice of a broader
TherapeuticsBench effort across drug-discovery stages and therapeutic
modalities. TxBench-PP tests whether agents can recover accurate conclusions
from real-world assay data rather than memorized facts from literature. The
benchmark contains 100 evaluations indexed by program stage, assay type, and
task structure, spanning mechanism-of-action (MoA) and pharmacodynamic (PD)
reasoning, compound-target engagement, causal target validation,
developability and safety, and translational efficacy. Agents receive realistic
workflow snapshots, inspect files in a coding environment, and return
structured answers graded deterministically. Across 16 model--harness
configurations, comprising 11 models and 4,800 trajectories, no system reliably
recovered preclinical
pharmacology decisions. The strongest configuration, Claude Opus 4.8 / Pi,
passed 59.3\% of endpoint attempts (178/300; 95\% CI, 51.1--67.6), followed by
GPT-5.5 / Pi at 55.3\% (166/300; 47.0--63.6).
\end{PreprintAbstract}

\vfill
\clearpage

\section*{TxBench-PP: Overview}

{\fontsize{9.1}{12.8}\selectfont
TxBench-PP is a focused TherapeuticsBench benchmark for small-molecule
preclinical pharmacology. Each evaluation supplies realistic workflow data
from a local preclinical decision point and asks whether an agent can recover
the result supported by the provided data.

Across 16 model--harness configurations, the strongest agent system passed
59.3\% of endpoint attempts (178/300; 95\% CI, 51.1--67.6).

TherapeuticsBench is the broader roadmap across drug-discovery stages and
therapeutic modalities. The schematic below contextualizes TxBench-PP with
future work.
\par}

\begin{center}
  \includegraphics[width=0.96\textwidth]{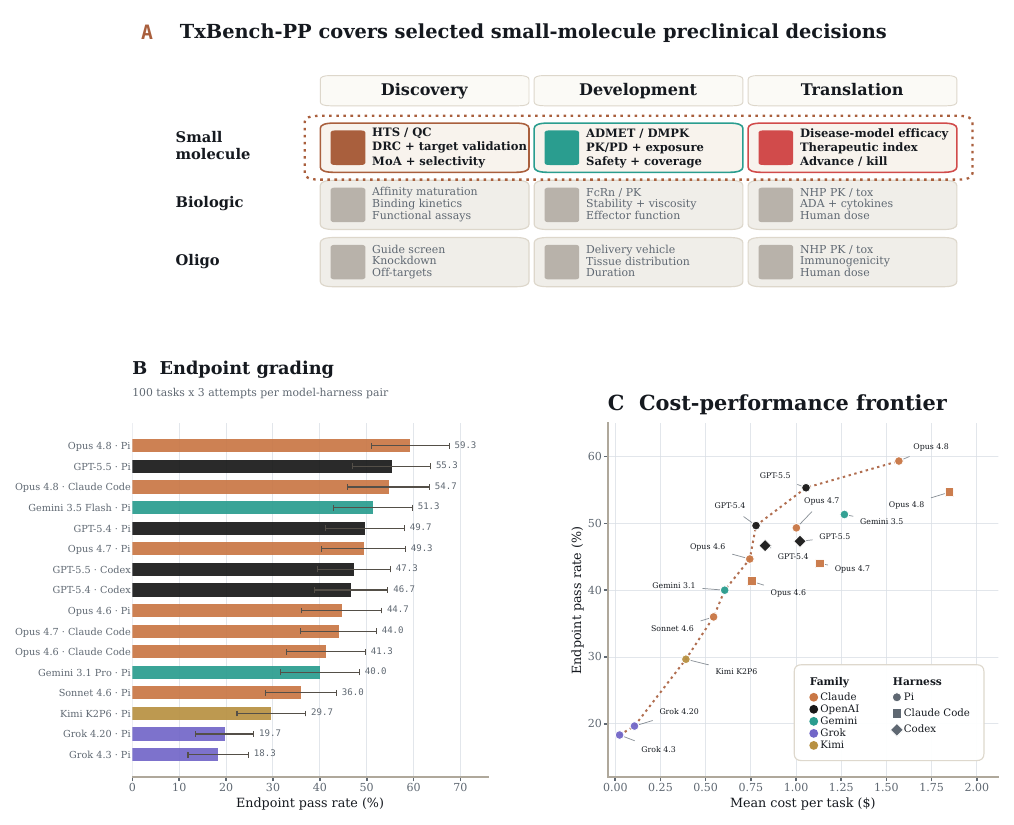}

  \captionof{figure}{\textbf{TxBench-PP scope and topline performance.}
  \textbf{A}, TxBench-PP places local small-molecule preclinical pharmacology
  decisions within the broader TherapeuticsBench roadmap. The dotted boundary
  marks the task groups in this release; biologics, oligonucleotides, and full
  program histories are out of scope. \textbf{B}, Endpoint pass rates across 16
  model--harness configurations; y-axis labels identify both model and harness.
  Confidence intervals are computed over task-level pass rates. \textbf{C}, Mean
  cost per task against endpoint pass rate; color denotes model family and
  marker shape denotes harness.}
  \label{fig:pipeline-scope}
\end{center}

\clearpage

{\fontsize{8.85}{12.0}\selectfont
\newcommand{\TwoColParaBreak}{\par\vspace{0.62em}\noindent}
\noindent\begin{minipage}[t]{0.47\textwidth}
\setlength{\parskip}{0pt}
\setlength{\parindent}{0pt}

\section{Introduction}

While experiments are rate-limited by natural processes, human decisions and
organizational consensus often make up significant components of program
timelines in drug discovery \cite{scannell2012,kola2004,cook2014,hughes2011}.
Agents promise to accelerate discovery, development, and translation by
compressing these interpretation and decision-making loops.  Specific knowledge
from constituent steps of the drug discovery pipeline can be encoded into
frontier models through post-training, tool use, and domain-specific
scaffolding, then deployed into practical data interpretation workflows
\cite{tuntland2014,vamathevan2019,hasselgren2024}.

\TwoColParaBreak However, the practical use of agentic systems in industrial
workflows requires standard and trusted methods of evaluating performance.
Recent biology-agent benchmarks have emphasized verifiable analysis tasks and
realistic scientific data
\cite{spatialbenchlong2026,spatialbench2025,scbench2026,labbench2024,bixbench2025,biomnibench2026,genebench2026}.
This is especially challenging in drug discovery because the ecosystem is a
sprawling landscape of assay categories, development stages, therapeutic
modalities, and decision types. Benchmarks must therefore measure realistic
tasks while providing focused treatment of the many local scientific judgements
that tile the biotech ecosystem.

\TwoColParaBreak

Preclinical small-molecule pharmacology is one important piece of this
landscape. Potency, target engagement, mechanism-of-action evidence,
pharmacodynamic response, exposure, safety, and in vivo efficacy are converted
into practical decisions about whether to advance, hold, repeat, or kill a
molecular candidate. These decisions sit upstream of costly medicinal
chemistry, animal studies, and human experiments. They are easy to get wrong
and often depend on fragmented expert judgement.

\TwoColParaBreak
We introduce TxBench-PP, a verifiable benchmark for evaluating AI agents on
small-molecule preclinical pharmacology. TxBench-PP is best understood as a
focused benchmark within the broader TherapeuticsBench roadmap rather than a
holistic evaluation of drug discovery. Each evaluation supplies
realistic workflow artifacts from a local preclinical decision point and asks
the agent to return a structured, deterministically gradable answer.
Importantly, the benchmark tests if agents can recover decisions from supplied
data instead of leaning on well-understood mechanisms and literature knowledge,
with many tasks intentionally designed to penalize memorized solutions.

\TwoColParaBreak
TxBench-PP contains 100 evaluations indexed by program stage, assay type, and
task structure. Tasks span mechanism-of-action and pharmacodynamic reasoning,
compound-target engagement, causal target validation, developability and safety,
and translational efficacy. Other parts of the therapeutic ecosystem, including
clinical program stages and non-small-molecule modalities, are left to future
TherapeuticsBench efforts. The benchmark offers a measuring stick by which
scientists can evaluate agentic systems on local preclinical pharmacology
decisions toward practical deployment in drug-discovery programs.

\end{minipage}\hfill
\begin{minipage}[t]{0.47\textwidth}
\setlength{\parskip}{0pt}
\setlength{\parindent}{0pt}

\section{Benchmark Construction}

Agents are evaluated on realistic preclinical pharmacology decisions over 100
evaluations spanning eight program stages: disease and model context, screening
and hit confirmation, drug response/pharmacogenomics, causal target validation,
MoA/PD, compound-target characterization, developability/safety/pharmacokinetics
(PK), and translational efficacy. Human genetic target-support tasks are
reserved for future benchmarking work. Across the benchmark, evaluations draw on assay types
including drug-response screens, genetic perturbation, transcriptomic and
proteomic profiling, chemoproteomic and live-cell target engagement, imaging,
drug metabolism and pharmacokinetics (DMPK) and absorption, distribution,
metabolism, excretion, and toxicity (ADMET) panels, toxicogenomics, and in vivo
efficacy or safety studies. These assay families build on established
measurement strategies for pooled drug response, growth-rate correction,
genetic dependency, transcriptional profiling, proteomics, target engagement,
toxicogenomics, and safety pharmacology
\cite{yu2016,corsello2020,hafner2016,meyers2017,pacini2021,subramanian2017,thompson2003,gerritsen2021,bantscheff2007,klaeger2017,martinezmolina2013,savitski2014,vasta2018,huang2016,igarashi2015,redfern2003,gintant2016}.

\TwoColParaBreak
The benchmark is built around recurring local decisions in small-molecule
programs. Candidate tasks begin as specific analysis points where assay data
might be turned into decisions. Examples include identifying a pathway mechanism
from dose-resolved phosphoproteomics, separating true chemoproteomic target
engagement from complex contaminants, and interpreting in vivo safety data.

\TwoColParaBreak
Each evaluation includes the data artifacts needed for that decision, metadata
needed to interpret the files, a task description, and a structured answer.
Task context is calibrated to approximate what a scientist would know at the
point of analysis. We avoid prescribing a method unless the method is part of
the scientific decision. The intended answer is a result supported by the
supplied data, not some fact or mechanism well understood in the literature.
Many tasks are designed to intentionally trap systems that overfit on trained
knowledge without empirical exploration.

\TwoColParaBreak
Candidate evaluations are refined through manual review and model trajectory
inspection. We remove tasks if the answer can be recovered by an obvious shortcut
that does not require interaction with data, when the prompt over-specifies the
analysis, or when plausible analysis choices produce answers that break grading
assumptions. The retained tasks are tagged by program stage, assay type, and
task structure.

\TwoColParaBreak
Grading uses deterministic functions over structured final answers. Depending
on the task, graders check label sets, numerical tolerances, rank-order
relationships, categorical choices, or all required fields. Manual trajectory
inspection is part of benchmark construction, following the same broad
verifiable-benchmark principle used in related biology-agent evaluations
\cite{spatialbenchlong2026,spatialbench2025,scbench2026}. We track
decision-blocking gaps that should change the program decision if recognized:
tox-species mismatch, insufficient free-drug coverage, hook effects,
contaminated chemoproteomic signals, survivor bias in pathology sampling, and
related failure points. These annotations are not exposed to agents; they are
used to interpret failures after endpoint grading and to guide future benchmark
updates.

\end{minipage}
}

\clearpage

{\fontsize{9.1}{12.8}\selectfont
\section{Benchmark Anatomy}

\begin{multicols}{2}
TxBench-PP is organized around the anatomy of a practical preclinical
pharmacology decision. Each evaluation is tagged along three axes: where the
decision sits in a small-molecule program, the assay type that generated the
data, and the structure of the task.
\par

The program-stage axis follows the labels shown in Figure~\ref{fig:benchmark-anatomy}:
disease and model context, screening and hit confirmation,
drug response/pharmacogenomics, human genetic target support, causal target
validation, MoA/PD, compound-target characterization,
developability/safety/PK, and translational efficacy. Human genetic target
support is reserved for future benchmarking work because adequately measuring
this category requires a separate focused benchmark. The order gives readers
context for task progression; it does not imply that every small-molecule
program proceeds through this strict sequence in practice
\cite{hughes2011,cook2014,nelson2015,minikel2024}.
\par

The assay and task axes describe the actual work required. Assay labels collapse
source kits into measurement families such as drug-response screens, protein or
post-translational modification (PTM) measurements, genetic perturbation, target
engagement, molecular state, safety profiling, and in vivo experiments. Task
labels describe broad families of decisions: mechanism inference, quality
control (QC), dependency calls, differential response, safety assessment,
imaging, hit prioritization, and related judgements.
\par
\end{multicols}
}

\begin{center}
  \includegraphics[width=0.96\textwidth]{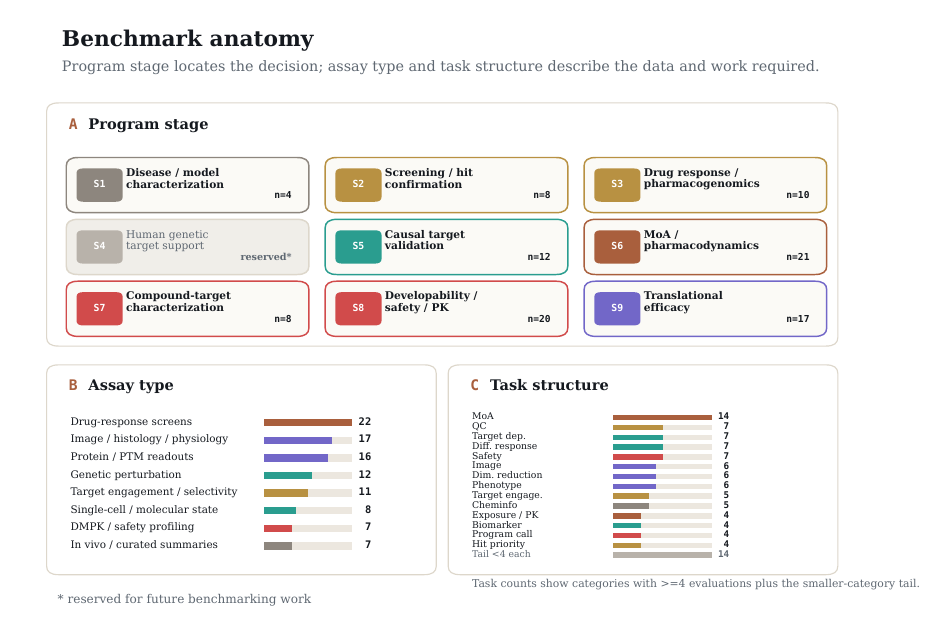}

  \captionof{figure}{\textbf{Benchmark anatomy.}
  \textbf{A}, Program stage labels place evaluations along a roughly ordered
  small-molecule evidence spine; the asterisk marks a stage reserved for future
  benchmarking work. \textbf{B}, Assay-type labels summarize the measurement
  families represented in the source data. \textbf{C}, Task-structure labels
  describe the operation or judgement required for a passing answer.}
  \label{fig:benchmark-anatomy}
\end{center}

\StartBody

\section{Results}

\subsection{Top systems remain below 60\%}

We evaluated 16 model--harness configurations, comprising 11 models across
three agent harnesses, on 100 preclinical pharmacology tasks. Each
configuration was run three independent times per task, yielding 4{,}800 agent
trajectories. A trajectory passed only when its submitted structured answer
satisfied every task-specific grader. We report endpoint pass rate as the
fraction of passing trajectories, with 95\% confidence intervals computed
across tasks.

The strongest configuration was Claude Opus 4.8 / Pi at 59.3\% (178/300
trajectories; 95\% CI, 51.1--67.6), followed by GPT-5.5 / Pi at 55.3\%
(166/300; 47.0--63.6), Claude Opus 4.8 / Claude Code at 54.7\% (164/300;
45.9--63.4), and Gemini 3.5 Flash / Pi at 51.3\% (154/300; 42.9--59.8).
These confidence intervals overlap, so the benchmark identifies a leading
group rather than a single clear winner. Even the top configuration failed 122
of 300 attempts and passed only 41 of 100 tasks in all three replicates.
Current agents therefore remain unreliable on local preclinical pharmacology
decisions, despite substantial differences in model family and harness
implementation.

\EndBody

\begin{center}
  \includegraphics[width=0.96\textwidth]{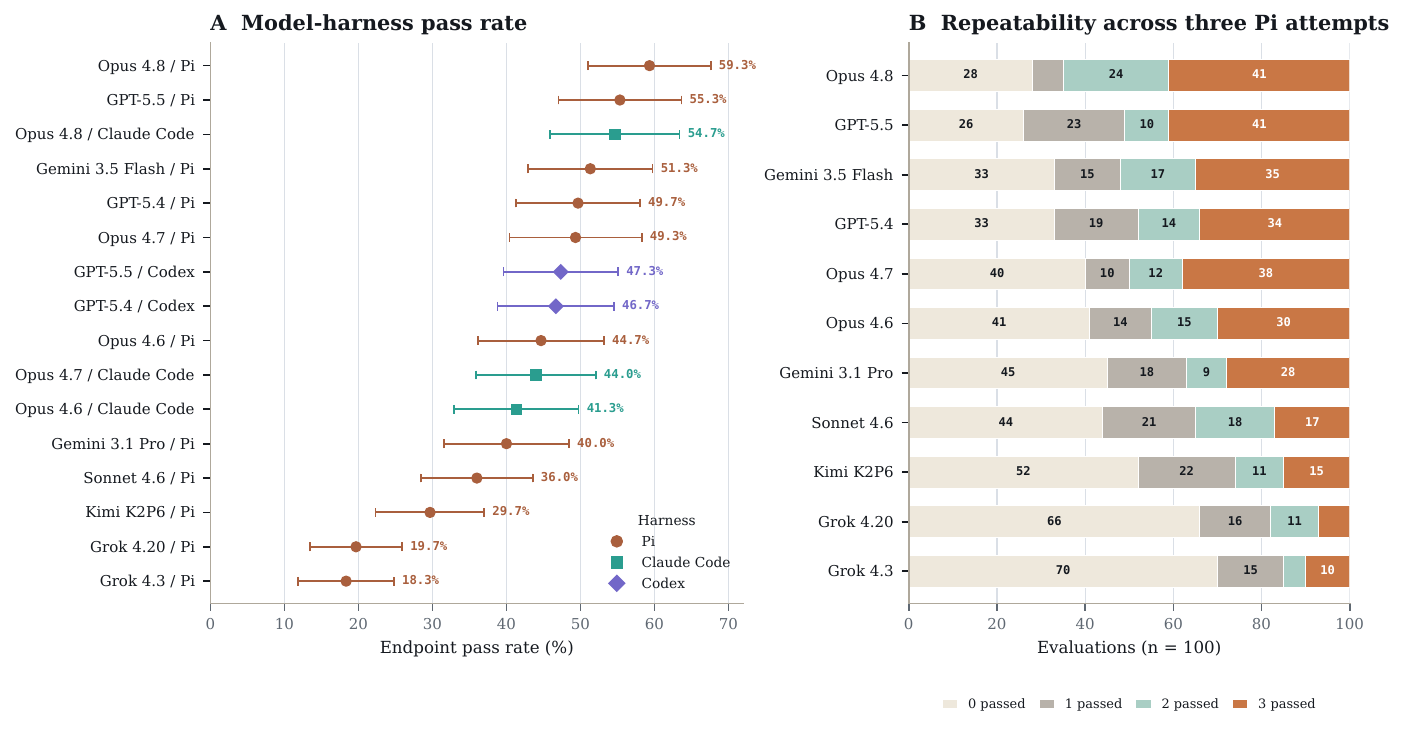}

  \captionof{figure}{\textbf{Top systems remain below 60\%.}
  \textbf{(A)} Endpoint pass rate for each of the 16 model--harness
  configurations across all 4{,}800 agent trajectories, with 95\% confidence
  intervals computed over task-level pass rates. Rows are sorted by pass rate;
  color and marker shape identify the agent harness. \textbf{(B)} Task-level
  reliability for Pi-harness models, partitioning the 100 evaluations by how
  many of the three independent rounds passed.}
  \label{fig:leaderboard}
\end{center}

\begin{center}
    \begin{minipage}{0.96\textwidth}
    \centering
    \fontsize{8.1}{9.3}\selectfont
    \setlength{\tabcolsep}{3.8pt}
    \renewcommand{\arraystretch}{0.9}
    \begin{tabular}{@{}lcccc@{}}
    \toprule
    Model / harness & Endpoint pass rate & $\geq 1/3$ & $\geq 2/3$ & $3/3$ \\
    \midrule
    Claude Opus 4.8 / Pi & 59.3\% (178/300; 51.1--67.6) & 72/100 & 65/100 & 41/100 \\
    GPT-5.5 / Pi & 55.3\% (166/300; 47.0--63.6) & 74/100 & 51/100 & 41/100 \\
    Claude Opus 4.8 / Claude Code & 54.7\% (164/300; 45.9--63.4) & 65/100 & 59/100 & 40/100 \\
    Gemini 3.5 Flash / Pi & 51.3\% (154/300; 42.9--59.8) & 67/100 & 52/100 & 35/100 \\
    GPT-5.4 / Pi & 49.7\% (149/300; 41.3--58.0) & 67/100 & 48/100 & 34/100 \\
    Claude Opus 4.7 / Pi & 49.3\% (148/300; 40.4--58.3) & 60/100 & 50/100 & 38/100 \\
    GPT-5.5 / OpenAI Codex & 47.3\% (142/300; 39.6--55.1) & 70/100 & 46/100 & 26/100 \\
    GPT-5.4 / OpenAI Codex & 46.7\% (140/300; 38.8--54.5) & 68/100 & 46/100 & 26/100 \\
    Claude Opus 4.6 / Pi & 44.7\% (134/300; 36.1--53.2) & 59/100 & 45/100 & 30/100 \\
    Claude Opus 4.7 / Claude Code & 44.0\% (132/300; 35.9--52.1) & 62/100 & 45/100 & 25/100 \\
    Claude Opus 4.6 / Claude Code & 41.3\% (124/300; 32.9--49.7) & 55/100 & 43/100 & 26/100 \\
    Gemini 3.1 Pro / Pi & 40.0\% (120/300; 31.5--48.5) & 55/100 & 37/100 & 28/100 \\
    Claude Sonnet 4.6 / Pi & 36.0\% (108/300; 28.4--43.6) & 56/100 & 35/100 & 17/100 \\
    Kimi K2P6 / Pi & 29.7\% (89/300; 22.3--37.0) & 48/100 & 26/100 & 15/100 \\
    Grok 4.20 reasoning / Pi & 19.7\% (59/300; 13.4--25.9) & 34/100 & 18/100 & 7/100 \\
    Grok 4.3 / Pi & 18.3\% (55/300; 11.9--24.8) & 30/100 & 15/100 & 10/100 \\
    \bottomrule
    \end{tabular}
    \captionof{table}{\textbf{Pass rates by model--harness configuration.}
    Pass rates are reported as percentages, with the numerator and denominator
    shown as passing trajectories out of 300 total trajectories per
    configuration, together with 95\% confidence intervals. Confidence intervals
    were computed across task-level mean pass rates over the 100 preclinical
    pharmacology tasks. The final three columns report the number of tasks, out
    of 100, for which at least one, at least two, or all three independent
    attempts passed.}
    \label{tab:passrates}
    \end{minipage}
\end{center}

\ResumeBody

\subsection{Trajectory analysis reveals gaps in scientific judgement}

We manually reviewed 1{,}834 failing Pi-harness trajectories. Most failures
revealed legitimate gaps in scientific judgement, where models inspected data
and performed plausible analyses but ultimately reached incorrect conclusions.

Among failures with an assignable scientific error, method and calibration
errors accounted for 71\%. Method errors included skipped or inappropriately
applied QC, incorrect statistical choices, and the use of an incorrect assay
data layer. In one CRISPR-screen task, a model aggregated guides to gene-level
scores without first removing guides that mapped to multiple loci, allowing
paralog-family artifacts to flood the essential-gene list. In a
proteasome-kinetics task, the supported answer lived in the ubiquitin-remnant
layer of a dual-readout mass-spectrometry run, but the agent incorrectly used
phosphoproteome data. Calibration errors came from cutoffs that were too
permissive or too strict for the decision. In one
competition-binding panel, models carried forward all 33 nonzero kinases when
only the 8 strong binders above a clear affinity gap were supported.

Other failures reflected incorrect image interpretation, memorized literature,
or claims not supported by the data. One model computed the strongest
data-supported biomarker, a vanadium compound correlated with SLC26A2 at
$r=-0.48$, then discarded it for the textbook elesclomol--FDX1 cuproptosis pair.
Another reported tumor regression from a sub-baseline median even though the
per-animal measurements showed growth arrest rather than regression.

Low step count was concentrated in the two weakest models, at 32\% of Grok
4.3's failures and 17\% of failures from Grok 4.20 reasoning, against 0--1\%
for every other model. Above that floor, more effort did not reliably improve accuracy.
Gemini 3.5 Flash took the most steps without leading the benchmark, while
GPT-5.5 improved on GPT-5.4 with fewer steps. Stronger models usually failed
after meaningful tool use.

\EndBody

\begin{center}
  \includegraphics[width=0.94\textwidth]{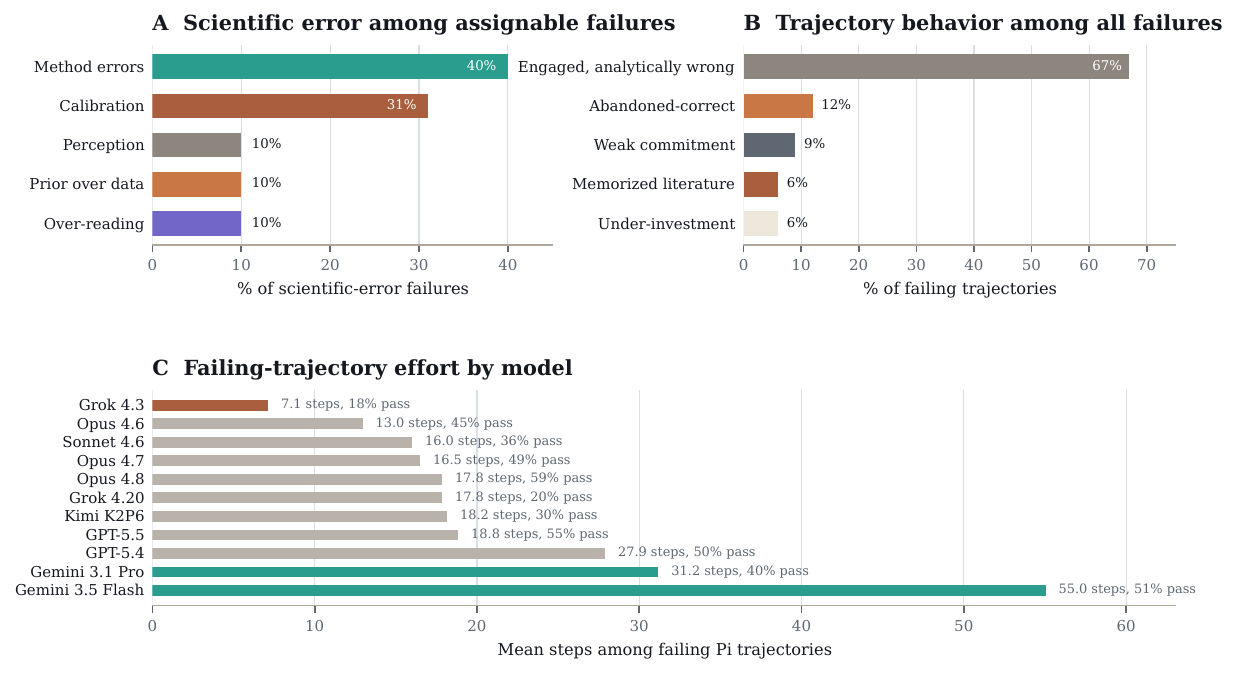}

  \captionof{figure}{\textbf{Reviewed failures mostly reflect scientific
  judgement errors.} The 1{,}834 reviewed failing Pi-harness trajectories were
  labeled by scientific error and trajectory behavior. \textbf{(A)} Among
  failures with an assignable scientific error, method and calibration errors
  dominate. \textbf{(B)} Most failures followed data engagement and plausible
  analysis; smaller shares came from abandoned-correct answers, weak
  commitment, memorized literature, or under-investment. \textbf{(C)} Mean step
  count among failing Pi-harness trajectories by model, with each model's
  endpoint pass rate shown next to the bar.}
  \label{fig:failure-taxonomy}
\end{center}

\ResumeBody

\subsection{Hard tasks require calibrated choices among candidates}

Performance varied by program stage, from 27\% in screening and hit
prioritization to 55\% in drug response (Figure~\ref{fig:by-stage}).
Difficult program stages involved decisions across QC, statistics, and chemical
or biological judgement of molecular candidates.

Analysis of trajectories helped diagnose model behavior. In screening, models
handled mechanical steps such as collapsing noisy replicates (\texttt{PRISM05},
58\%), but were unable to apply biological judgement to separate technical
artifacts from plausible leads. On outlier-pool QC (\texttt{PRISM01}, 18\%),
models excluded many cytotoxic positive-control wells as artifacts, and on
oncology drug repurposing (\texttt{PRISM\_S2\_11}, 0\%) they ranked by breadth
of killing rather than selectivity. In causal target validation, models
recovered clean dependencies but skipped guide-level QC: on
multi-target guide contamination (\texttt{GG\_01}, 16\%) they aggregated guides
without removing multi-locus single-guide RNAs (sgRNAs). In target engagement,
models detected binding signals but lacked a stable cutoff, accepting every
weak binder in one kinase-binding panel (\texttt{TXBX\_KB\_v11\_DA1}, 0\%) and
discarding genuinely engaged targets behind a fabricated stringency cutoff in
another (\texttt{CTRL01}, 21\%). In translational efficacy, models consistently
struggled to reason accurately about microscopy or organoid images: on a
two-marker immunostain (\texttt{HAI2027\_MM\_FIG2E}, 42\%) models credited
faint, near-vehicle single-agent staining as activity, and in organoid
viability tasks apoptotic debris was sometimes counted as living organoids.

Task-type aggregation shows the same structure. Across the 14 task categories
with at least four evaluations, the mean Pi-harness pass rate across the 11
models ranged from 70\% on exposure/PK tasks to 19\% on hit prioritization and
17\% on cheminformatics (Figure~\ref{fig:by-task-type}). Exposure/PK,
biomarker discovery, safety assessment, and target engagement averaged 57\%;
these tasks usually ask for a computed value, a binary call, or one supported
entity. Target dependency, differential response, hit prioritization, and
cheminformatics averaged 24\%; these require selecting a defensible subset from
many plausible candidates. No model exceeded 45\% on hit prioritization or
cheminformatics, indicating a shared limit across current systems.

\EndBody

\begin{center}
  \includegraphics[width=0.94\textwidth]{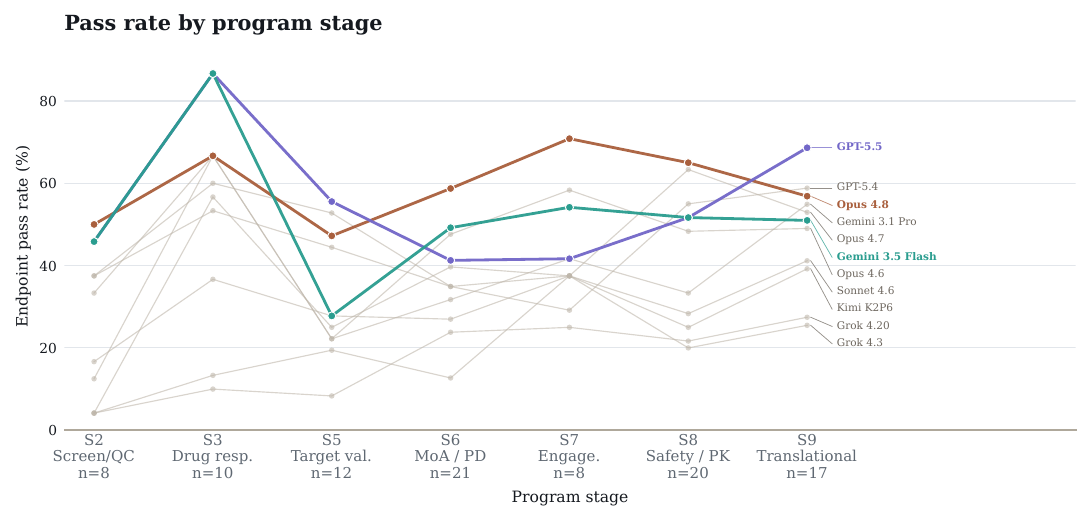}

  \captionof{figure}{\textbf{Pass rate by program stage.} Endpoint pass rate by
  program stage across Pi-harness models. S1 disease/model characterization is
  omitted because it is backed by too few evaluations; S4 human genetic target
  support is reserved for future benchmarking work.}
  \label{fig:by-stage}
\end{center}

\begin{center}
  \includegraphics[width=0.94\textwidth]{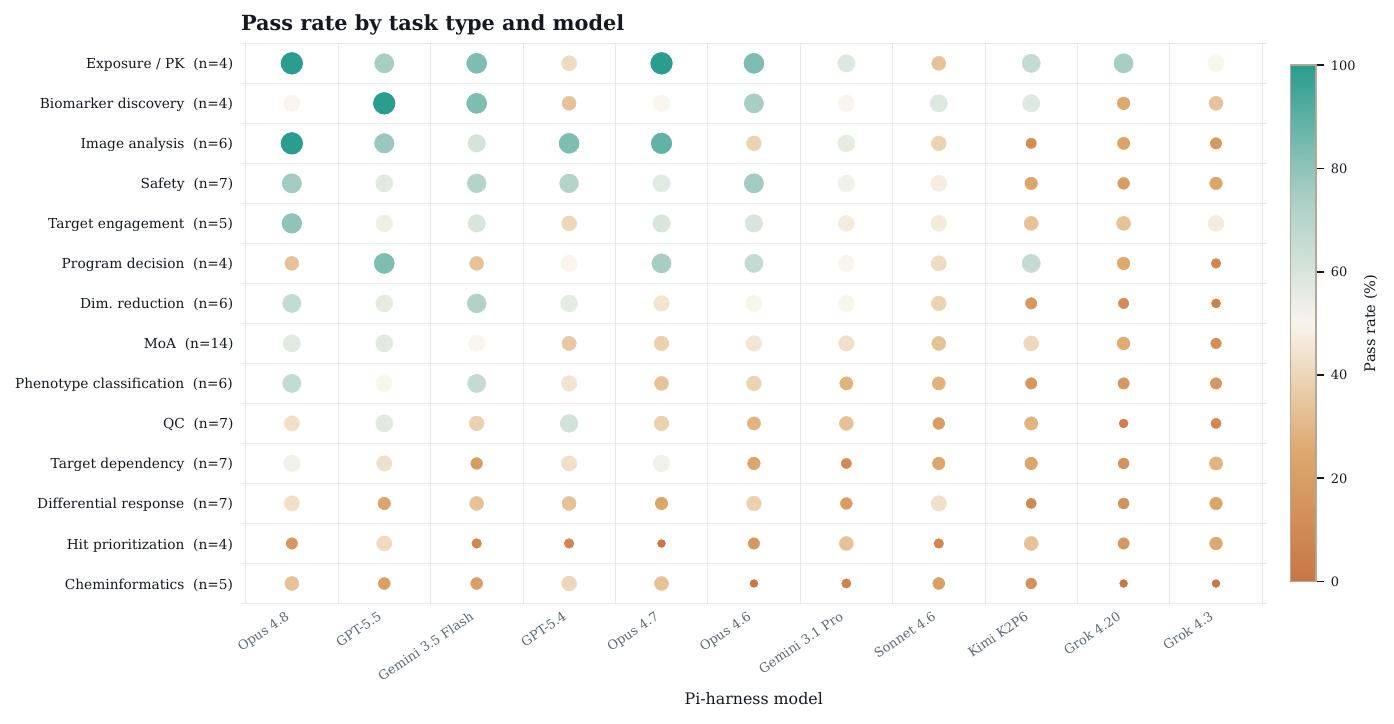}

  \captionof{figure}{\textbf{Task-type performance separates determinate calls
  from candidate selection.} Pi-harness pass rate by task type and model for
  categories with at least four evaluations. Rows are ordered by the mean across
  the 11 models.}
  \label{fig:by-task-type}
\end{center}

\ResumeBody

\subsection{Program decisions expose practical failures}

Some evaluations closely resemble real program decisions, giving the agent
several assay readouts at once and asking for a single advance, hold, or
prioritization call. The current suite includes seven such advancement
decisions; across 230 runs, models pass only 35\% of them. Models perform well
when the data are clean and consistent, but struggle on more challenging cases,
either advancing weak candidates or discarding strong ones.

\emph{Clean correct.} An example organoid program go/no-go task is the clean
case. Given a primary screen, a mechanism image, a toxicology counterscreen, and
DMPK exposure for eight blinded compounds, the correct action is to advance the
one compound that shows the desired cytostatic mechanism together with an
adequate safety margin and exposure coverage, and most models do (82\%).

\emph{Wrong synthesis.} On some decision tasks, the agent must assemble a set
of experimental models with convergent support across several endpoints, but
models incorrectly overfit on a single endpoint (27\%), so the program's next
experiments are built on wrong assumptions.

\emph{False-go.} An example safety advancement task requires that no candidate
move forward, because none pairs activity with healthy-tissue sparing, yet
models advance one on activity alone or on a relative tumor-versus-healthy index
(36\%), putting an unsafe compound into the pipeline.

\emph{False no-go.} The reverse error also appears. On an example long-term
combination task, models read a strong colony-stain reduction as a reason to
drop a candidate, incorrectly culling a program with strong data
(\texttt{HAI2027\_FIG1\_DUR01}, 48\%).

Strong overall benchmark scores did not reliably predict these program
decisions. The six advancement decisions with complete Pi-harness model
comparisons yielded an order almost unrelated to overall accuracy (Spearman
$\rho = 0.08$; Figure~\ref{fig:program-model-compare}A).
Claude Opus 4.8, first overall at 59.3\%, ranked last on these decisions at
22.2\% (4/18).

\EndBody

\begin{center}
  \includegraphics[width=0.96\textwidth]{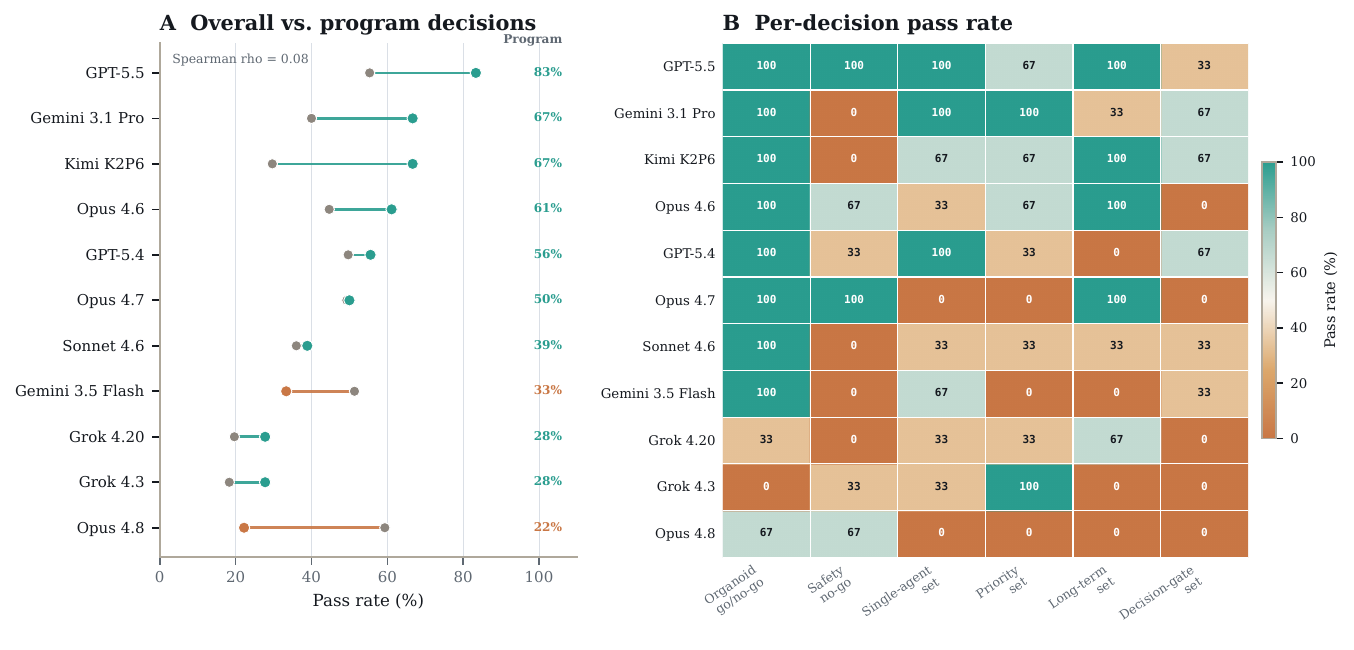}

  \captionof{figure}{\textbf{Overall accuracy does not predict advancement
  decisions.} Pass rates on the six advancement decisions with complete
  Pi-harness model comparisons (up to 18 runs per model). \textbf{(A)} Overall
  endpoint pass rate versus pass rate on the six program decisions, with rows
  sorted by program-decision performance (Spearman $\rho = 0.08$). \textbf{(B)}
  Per-decision pass rate by model. The two single-compound calls are passed
  broadly, but tasks that require selecting the supported model set separate the
  field.}
  \label{fig:program-model-compare}
\end{center}

\ResumeBody

\subsection{Scores depend on the agent harness}

Each harness used the same model checkpoint, task text, and answer-submission
protocol, but pass rate still varied by harness. In matched comparisons within
the same model family, Pi outperformed Claude Code by 4.4 percentage points
(95\% CI: 1.2--7.8) on the shared Claude Opus models and outperformed OpenAI
Codex by 5.5 percentage points (95\% CI: 1.2--9.8) on the shared GPT models. Pi
scored higher on every shared model, not only on average.

The size of this difference is comparable to improvements in model generations
within families. Moving from Claude Opus 4.6 to Claude Opus 4.7 improved Pi
performance by 4.6 points, similar to the harness effect. The Pi advantage was
also similar on image-based and non-image tasks, so it is not explained only by
differences in image reasoning. Claude Code rendered figures to the model in
100\% of image-bearing tasks, compared with 84\% for Pi; OpenAI Codex, which
lacks vision capabilities, never rendered figures.

The remaining differences point to implementation details. Pi's lean,
three-tool loop failed on 0.6\% of runs, compared with 1.8\% for Claude Code,
and reached answers in fewer steps. OpenAI Codex was also penalized on
figure-dependent tasks because it did not render figures. These effects are
small in absolute terms and have wide confidence intervals, but they show that
reported scores are properties of the model--harness configuration rather than
the model alone.

\EndBody

\begin{center}
  \begin{minipage}{0.96\textwidth}
  \centering
  \includegraphics[width=\textwidth]{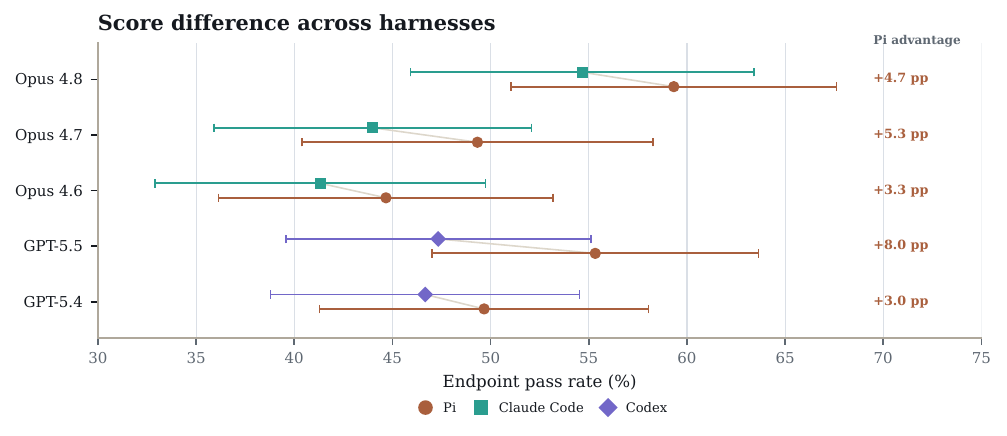}

  \captionof{figure}{\textbf{Harness choice changes scores within the same
  model.} Endpoint pass rates for models run under Pi and an alternate harness,
  with 95\% confidence intervals computed over task-level pass rates. Each row
  compares the same model checkpoint on the same task set; labels at right show
  the within-model Pi advantage in percentage points. These matched comparisons
  support interpreting reported scores as properties of the model--harness
  configuration, not the model alone.}
  \label{fig:harness-impact}
  \end{minipage}
\end{center}

\ResumeBody

\section{Discussion}

TxBench-PP measures a small but relevant category of drug discovery work:
whether an agent can make accurate scientific decisions about preclinical
pharmacology data with a similar amount of context to what a scientist would
have in practice. It is a small-molecule-focused benchmark within a broader
TherapeuticsBench roadmap, and should not be used to make more general claims
about other areas of the ecosystem, such as human genetic target support, full
disease-model selection, broad clinical translation, or non-small-molecule
modalities.

The results show agents can complete some tasks correctly but cannot yet be
trusted as reliable scientific assistants. Claude Opus 4.8 / Pi was the leading
configuration, passing 59.3\% of endpoint attempts (178/300; 95\% CI,
51.1--67.6), but still failed more than two of every five attempts. Manual
trajectory review revealed specific classes of scientific failure. Agents
engaged in productive, exploratory behavior but made errors reasoning about QC,
statistics, biological context, or molecular properties, or overfit on
memorized facts. Local analysis errors can advance weak or unsafe candidates and
discard supported ones. The data suggest work remains before these systems can
be integrated reliably and reproducibly into preclinical biology workflows.

Future work will extend the benchmark downstream to clinical tasks, vertically
to broader pharmacogenomics, and
across modalities to antibodies, antibody-drug conjugates (ADCs), degraders,
oligonucleotides, cell therapies, and gene therapies. Complementary long-horizon
TherapeuticsBench work will follow entire programs across discovery,
development, and translation, analogous to long-horizon benchmark designs in
other biological domains \cite{spatialbenchlong2026}.

\columnbreak

\section{Methods}

\subsection{Benchmark assembly}

Benchmark construction is described in the Benchmark Construction and Benchmark
Anatomy sections. Briefly, we assembled 100 evaluations from practical
small-molecule preclinical pharmacology workflow states, using construction
principles adapted from prior verifiable biology-agent benchmarks
\cite{spatialbenchlong2026,spatialbench2025,scbench2026}. Candidate
evaluations were retained when the target decision could be recovered from the
supplied data artifacts, expressed through a constrained final answer, and
graded deterministically. Tasks were revised or removed when the prompt
over-specified the method, when the answer could be recovered by an obvious
shortcut, when plausible analysis choices produced incompatible decisions, or
when the grader could not distinguish the supported pharmacology conclusion
from a plausible but unsupported answer.

Each retained evaluation includes a task prompt, workflow data artifacts,
metadata needed to interpret the files, a structured answer schema, a hidden
deterministic grader, and tags for program stage, assay type, and task
structure. Evaluation notes record the intended empirical decision, known
analysis traps, and decision-blocking gaps used for later trajectory
interpretation. These notes and failure annotations are not shown to agents.
Safety, exposure, toxicogenomics, and cheminformatics evaluations used standard
concepts from cardiac safety, free-drug and tissue-exposure reasoning,
toxicogenomics resources, molecular fingerprints, and scaffold analysis
\cite{redfern2003,ichs7b2005,gintant2016,webborn2025,zhang2019,huang2016,igarashi2015,rogers2010,bemis1996}.

\subsection{Agent runs}

We evaluated 16 model--harness configurations across Pi, Claude Code, and
OpenAI Codex. Pi was run with Claude Opus 4.6, Claude Opus 4.7, Claude Opus
4.8, Claude Sonnet 4.6, GPT-5.4, GPT-5.5, Gemini 3.1 Pro, Gemini 3.5 Flash,
Kimi K2P6, Grok 4.20 reasoning, and Grok 4.3. Claude Code was run with Claude
Opus 4.6, Claude Opus 4.7, and Claude Opus 4.8. OpenAI Codex was run with
GPT-5.4 and GPT-5.5.

Each configuration was run on the same 100 evaluations with three independent
attempts per evaluation, yielding 300 trajectories per configuration and 4,800
trajectories overall. Runs used the packaged task context, the supplied data
artifacts, a coding environment for file inspection and analysis, and the same
structured answer-submission protocol. For each run we recorded the final
answer, grader output, trajectory, result log, duration, step count, and
available cost metadata.

Failed, timed-out, incomplete, malformed, or unparsable runs were retained in
the denominator and counted as nonpassing unless the hidden grader emitted a
literal boolean passing result. The result parser treated missing pass fields
or non-boolean pass fields as failures.

\subsection{Endpoint grading}

The primary benchmark score is deterministic endpoint grading of the final
structured answer. Graders parse the submitted answer and compare the requested
fields against task-specific criteria. Depending on the evaluation, these
criteria include exact or normalized label matches, numerical tolerances,
rank-order checks, categorical choices, and all-of field comparisons. A run
passes only when all required task-specific checks pass. Missing fields,
invalid JSON, answers outside the expected schema, and answers that cannot be
parsed by the grader are counted as failures.

We also report task-level robustness counts for each model--harness
configuration: the number of evaluations for which at least one, at least two,
or all three replicate attempts passed. These counts are diagnostic and do not
replace the endpoint pass rate.

\subsection{Result aggregation and statistical analysis}

For each model--harness configuration and evaluation, the three replicate
passes were averaged to produce an evaluation-level pass score in
\{0, 1/3, 2/3, 1\}. The primary endpoint pass rate is the mean of these 100
evaluation-level scores. We report raw pass counts for interpretability, but
compute confidence intervals over evaluations rather than over individual
trajectories, because replicate attempts on the same evaluation are not
independent samples of the benchmark. Unless otherwise stated, 95\% confidence
intervals are Student $t$ intervals over evaluation-level mean pass scores.

Stage, assay, and task-type analyses use the same evaluation-level aggregation
within each metadata-defined subset. Task-type plots are restricted to
categories with at least four evaluations. Cost, duration, and step summaries
are averaged first across the three replicate attempts for each evaluation and
then across evaluations. Cost analyses use the recorded total-cost metadata
when available; for Anthropic model runs without a total-cost field, cost was
computed from recorded token usage and the same model-specific price table used
for aggregation.

Harness comparisons are matched by model and evaluation. The Claude-family
comparison uses Claude Opus 4.6, Claude Opus 4.7, and Claude Opus 4.8 on Pi
and Claude Code. The GPT-family comparison uses GPT-5.4 and GPT-5.5 on Pi and
OpenAI Codex. Program-decision analyses use the curated subset of evaluations
that ask for an advance, hold, prioritize, or related preclinical program
decision from multiple evidence streams; the model-comparison panel uses the
six such decisions with complete Pi-harness comparisons across the 11 Pi
models.

\subsection{Manual trajectory review}

Manual trajectory inspection was used during benchmark construction and for
post hoc failure analysis. During construction, trajectories were inspected to
identify prompts that over-specified the method, shortcut-solvable tasks,
unstable answer surfaces, and graders that failed to separate supported
answers from plausible but unsupported ones.

For the failure taxonomy, reviewers labeled 1{,}834 failing Pi-harness
trajectories against the evaluation's documented scientific intent. Each
reviewed failure received a behavioral-cause label describing how the model
failed. Failures with an assignable scientific issue also received a
scientific-error label describing what was wrong with the submitted answer.
These labels were used only for diagnostic analysis after endpoint grading.

\section*{Data availability}

The public TxBench-PP repository is available at
\url{https://github.com/latchbio/txbench-pp}. It contains a subset of
evaluations and trajectories for manual inspection, along with task definitions,
graders, and available data artifacts for those evaluations. Source assay data
are distributed or linked according to the original data licenses and access
terms.

{\fontsize{8.3}{10.7}\selectfont

}

\EndBody

\end{document}